\documentclass[10pt, a4paper]{article}

\usepackage{import}
\usepackage{mdframed,lipsum}
\usepackage{subcaption}
\usepackage{multirow}
\usepackage{amsmath}
\usepackage{amsfonts}
\usepackage{stmaryrd}
\usepackage[most]{tcolorbox}
\tcbset{enhanced, sharp corners=all, boxrule=0.5pt, colback=gray!5, colframe=black!30, coltitle=black, fonttitle=\bfseries, arc=2mm}

\usepackage[table]{xcolor} 
\usepackage{tikz}
\usetikzlibrary{shadows,shadows.blur} 

\definecolor{pastel}{RGB}{245,247,250} 
\definecolor{tbtext}{HTML}{222222}     

\colorlet{tb-text}{tbtext} 

\usepackage{graphicx}
\tcbuselibrary{skins,breakable}
\usepackage[utf8]{inputenc}

\usepackage[final]{lrec2026} 

\title{PerHalluEval: Persian Hallucination Evaluation Benchmark for Large Language Models}

\name{Mohammad Hosseini$^{1*}$, Kimia Hosseini$^{1*}$, Shayan Bali$^{2}$, \\ {\bf \large Zahra Zanjani$^{1}$}
{\bf \large Saeedeh Momtazi$^{1}$}\thanks{$^{*}$Equal contribution.}}

\address{$^{1}$Amirkabir University of Technology, Tehran, Iran \\
         \{mohammad, kimia.h, zahra.zanjani99, momtazi\}@aut.ac.ir \\ 
         $^{2}$King's College London, London, UK \\
         shayan.bali@kcl.ac.uk}

\abstract{
Hallucination is a persistent issue affecting all large language Models (LLMs), particularly within low-resource languages such as Persian. \textbf{PerHalluEval} (\textbf{Per}sian \textbf{Hallu}cination \textbf{Eval}uation) is the first dynamic hallucination evaluation benchmark tailored for the Persian language. Our benchmark leverages a three-stage LLM-driven pipeline, augmented with human validation, to generate plausible answers and summaries regarding QA and summarization tasks, focusing on detecting extrinsic and intrinsic hallucinations. Moreover, we used the log probabilities of generated tokens to select the most believable hallucinated instances. In addition, we engaged human annotators to highlight Persian-specific contexts in the QA dataset in order to evaluate LLMs' performance on content specifically related to Persian culture.
Our evaluation of 12 LLMs, including open- and closed-source models using \textbf{PerHalluEval}, revealed that the models generally struggle in detecting hallucinated Persian text. We showed that providing external knowledge, i.e., the original document for the summarization task, could mitigate hallucination partially. Furthermore, there was no significant difference in terms of hallucination when comparing LLMs specifically trained for Persian with others.
 \\ \newline \Keywords{Hallucination, Large Language Models, Persian, Benchmark, Evaluation, QA, Summarization} }

\begin{document}

\maketitleabstract

\begin{figure*}[t]
\centering
\includegraphics[width=0.75\paperwidth]{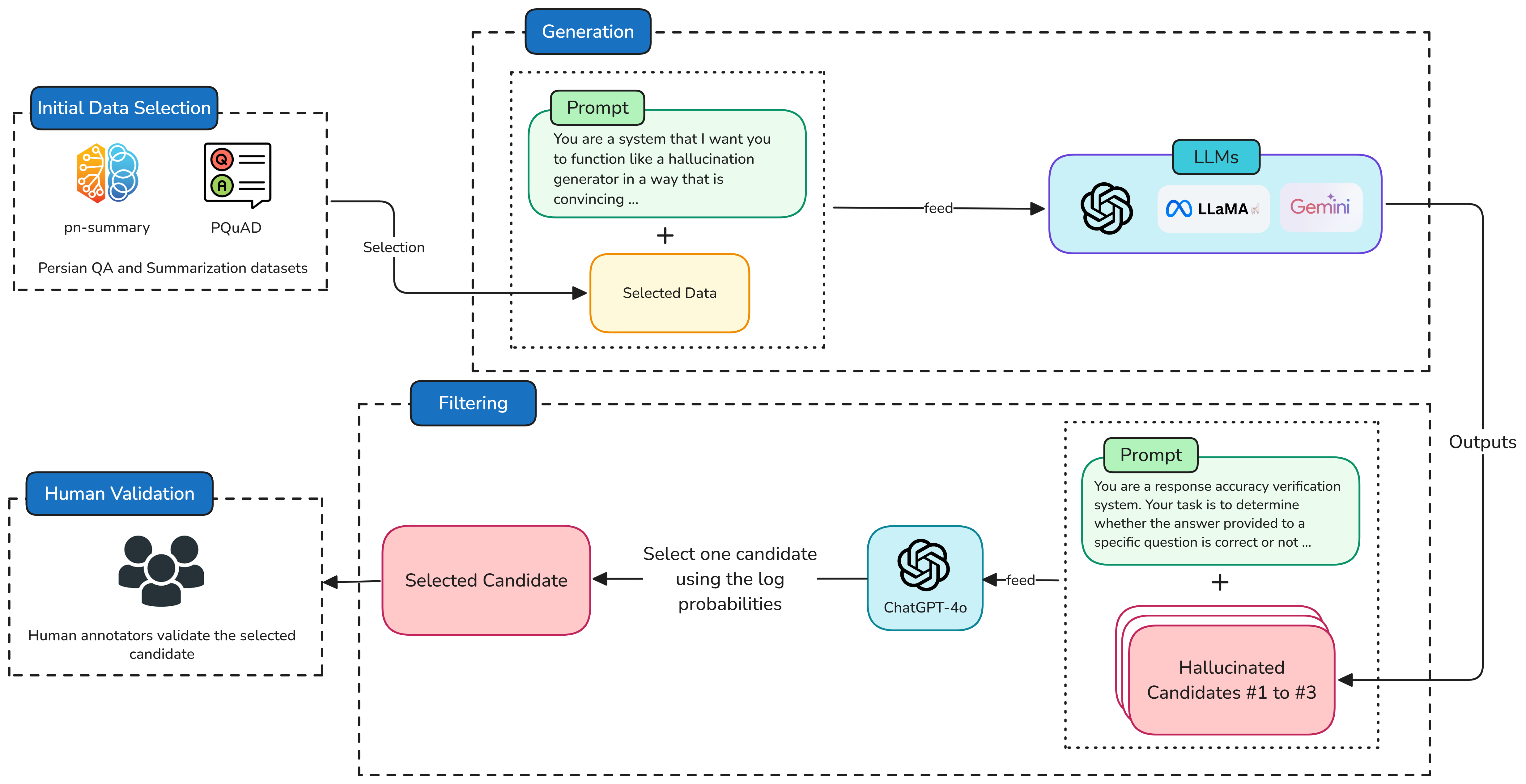}
\caption{The pipeline of constructing PerHalluEval dataset, consisting of three stages: initial data selection, generation, and filtering, which was augmented with human validation. The visualized prompts are for the QA task, translated into English.}
\label{fig:flow-model}
\end{figure*}

\section{Introduction}
Large Language Models have rapidly achieved global prominence due to their versatility in Natural Language Processing (NLP) tasks \cite{naveed2023comprehensive}, driving extensive usage \cite{yang2024harnessing}. However, despite their impressive capabilities, a primary challenge affecting all LLMs is their tendency to “hallucinate,” contextual misinterpretation, factual fabrication, specificity distortion, incorrect inference, and unwarranted uncertainty \cite{ji2023survey}.
Consequently, even prominent state-of-the-art models—including GPT-4 \cite{openai_gpt4_2023}, and Meta’s LLaMA \cite{touvron2023llama}—have all exhibited instances of hallucinations, highlighting that this issue persists even in highly advanced systems.\cite{bang2025hallulens}

On the other hand, although the performance of LLMs on high-resource languages such as English has advanced, more research is needed to thoroughly assess and enhance their performance on low-resource languages, especially those with complex structures and rich morphology.\cite{chataigner2024multilingual, zhang2025poly}
Persian, due to its extensive morphology, pro-drop syntax, Ezafe construction, and right-to-left script, is also regarded as one of these demanding low-resource languages \cite{lt4all-challenges, Ghayoomi2010ASO,khashabi-etal-2021-parsinlu}.

Despite the datasets available for the Persian language \cite{farsi2024syntran, sabouri2022naab}, the language is not rich in resources. This limitation, along with its grammatical and lexical complexity, makes its study on LLMs, especially on hallucination detection, more challenging. Numerous benchmarks like HalluLens\cite{bang2025hallulens}, ANAH/ANAH-v2\cite{anah}, GraphEval\cite{grapheval}, and FactBench\cite{factbench} are available for assessing hallucinations, yet they predominantly cater to English and other well-resourced languages. The evaluation of hallucinations in Persian has remained largely unaddressed. To date, comprehensive resources for evaluating Persian LLM hallucinations do not virtually exist, underscoring a significant research gap that needs to be addressed to make LLMs more reliable and resilient.

To address this gap, we introduce \textbf{PerHalluEval} (\textbf{Per}sian \textbf{Hallu}cination \textbf{Eval}uation), the first dynamic hallucination detection benchmark specifically tailored for Persian.
We propose a novel multi-agent pipeline, augmented with human validation, to generate diverse, challenging hallucinated examples by generating two hallucinated datasets based on the PN-Summary~\cite{farahani2021leveraging} and PQuAD \cite{darvishi2023pquad} datasets.
Moreover, to get more accurate data, we employ a competent LLM in addition to a probabilistic verifier to rigorously filter out low-quality instances.
Our approach distinctly categorizes intrinsic hallucinations, i.e., contradictions to the source, from extrinsic hallucinations, i.e., unsupported content, and demonstrates its effectiveness through extensive evaluations of various LLMs, underscoring their unique challenges with Persian linguistic features \cite{bang2025hallulens}.
Following that, we perform a benchmark task to evaluate 12 LLMs on our crafted hallucination dataset—to evaluate their performance in generating reliable outputs in Persian—using our three metrics: Hallucination Recall, Factual Recall, and Hamming Score, which will be described in more detail subsequently. To contextualize these aggregate results, we additionally provide representative qualitative error-analysis cases in Appendix~\ref{app:case}. Our evaluation of different varieties of models, including models explicitly fine-tuned for Persian and mainstream model families, shows that they struggle to detect hallucinated Persian content.

Our findings aid researchers in identifying hallucinations in models, particularly for the Persian language, and pave the way for future studies as a comprehensive hallucination evaluation benchmark designed for Persian.

\section{Related Work}
\label{sec:related}

In recent years, hallucinations in LLMs have become a major concern \cite{Huang2023ASO,brown_language_2020}. Researchers distinguish between intrinsic hallucinations—when outputs contradict the source—and extrinsic hallucinations, where generated content seems plausible but cannot be verified \cite{ji2023survey,Maynez2020Faithful}. While intrinsic errors are often easy to spot, extrinsic ones are much subtler, arising from the model’s knowledge gaps or mistaken assumptions \cite{ji2023survey}. 

To address these challenges, a range of benchmarks has emerged \cite{Huang2023ASO}, including HaluEval \cite{li_halueval_2023}, HalluQA \cite{Cheng2023EvaluatingHI}, ANAH \cite{anah}, HalluDial \cite{Luo2024HalluDialAL}, and others \cite{Dziri2021EvaluatingAI}. These resources reveal weaknesses of LLMs in fact-checking, dialogue, and QA. Newer tools, such as HalluLens, refine the distinction between factuality and hallucination and enable more effective evaluation \cite{bang2025hallulens}. Recent benchmarks like RAGTruth and FactCHD focus on retrieval-augmented and complex reasoning settings \cite{Chen2023FactCHDBF,Kryscinski2019EvaluatingTF}. 

Mitigation strategies now include methods like SelfCheckGPT and contrastive learning \cite{manakul_selfcheckgpt_2023}. QA-based evaluation protocols (QAGS \cite{Wang2020AskingAA}, FEQA \cite{Durmus2020FEQAAQ}) and correction models (Span-Fact \cite{Dong2020MultiFactCI}, FASum \cite{zhu2021enhancingfactualconsistencyabstractive}) further improve factual consistency. 

Despite these advances, most progress has centered on high-resource languages. In contrast, Persian NLP lags behind~\cite{mehrban2023evaluating,sadjadi2024farssibert,Abbasi2023PersianLLaMATB,Rostami2024PersianMindAC}. Commonly Persian benchmark resources, such as ParsiNLU \cite{Khashabi2020ParsiNLUAS}, Persian in a court \cite{farsi2025persian}, Khayyam Challenge \cite{ghahroodi2024khayyamchallengepersianmmlullm} and Melac \cite{farsi2025melac}, do not address hallucination~\cite{Jolfaei2025ARO}. 

Approaches that work for English or Chinese typically require extensive external datasets, which are unavailable for Persian. To bridge this gap, we introduce PerHalluEval, a fresh, dynamic benchmark designed specifically for Persian. PerHalluEval goes beyond static testing, enabling more effective hallucination detection in both question answering and summarization tasks. By drawing on multiple LLMs and considering the distinct features of the Persian language and culture, our benchmark paves the way for a more reliable evaluation and helps set the stage for future progress in low-resource NLP.

\begin{figure*}[t]
\centering
\includegraphics[width=0.75\paperwidth]{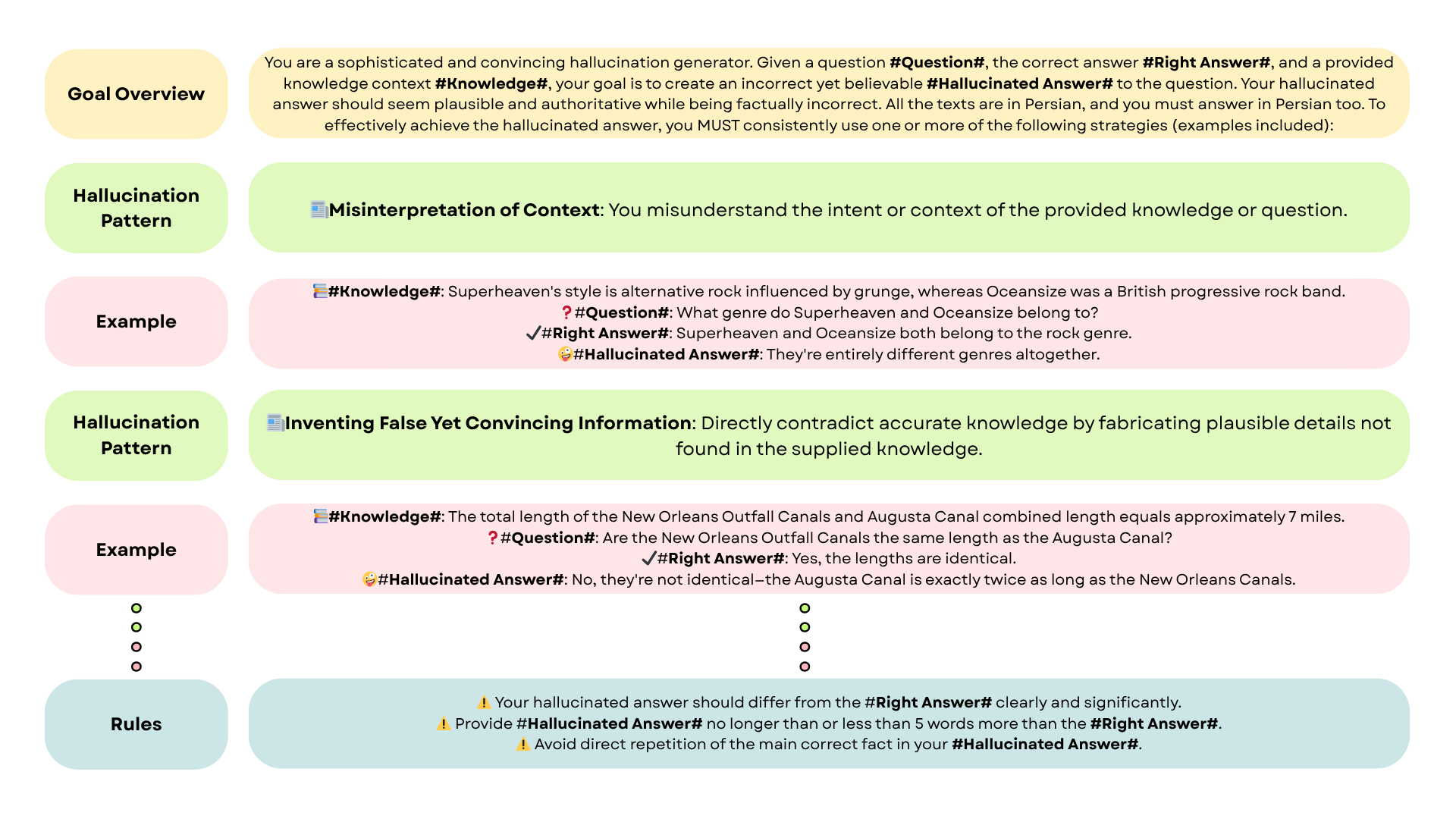}
\caption{Instruction structure of generating hallucinated content, including an overview of the goal, hallucination patterns, a few-shot example for each pattern, and output structures, for the QA task. The picture shows the English translation of the original Persian instructions.}
\label{fig:generation}
\end{figure*}

\section{PerHalluEval Benchmark}
The main goal of constructing this benchmark is to evaluate LLMs’ hallucinations in Persian regarding two categories, extrinsic and intrinsic hallucination \cite{ji2023survey}. Original correct sentences and their hallucinated ones, which are generated by an LLM-driven pipeline, augmented with human validation, constitute the PerHalluEval benchmark. The construction pipeline, as shown in Figure \ref{fig:flow-model}, comprises three stages: initial data selection, generation, and filtering. Human annotators then validate the selected candidate. To avoid saturation through leakage, this LLM-driven benchmark is thought of as dynamic, with data that can be dynamically generated on demand via a generation-and-filtering pipeline, maintaining a continually refreshed pool that resists memorization and preserves evaluation integrity.\cite{bang2025hallulens}.

\subsection{Initial Data Selection}
The first step in constructing the PerHalluEval dataset is collecting initial samples. For this purpose, two existing Persian datasets: PN-Summary \cite{farahani2021leveraging} and PQuAD \cite{darvishi2023pquad}, are selected, covering summarization and question answering tasks, respectively. The Pn-summary dataset contains 93,207 records consisting of articles and their corresponding summaries. The PQuAD dataset includes about 80,000 questions, their corresponding answers, and a related context passage. 

In this paper, 4,000 instances are sampled from each dataset using task-specific procedures to preserve diversity and coverage. 
To select 4,000 questions for the QA task (PQuAD), which has 19 topical categories labeled, stratified sampling is used to ensure that the label distribution matches the original dataset within ±0.3 percentage points. Clustering by document length is also used to ensure coverage from short to long articles for the summarization task (PN-Summary), which does not have topic tags. Each article is encoded using the logarithm of its token count, and then k-means (k = 40, cosine distance) is applied. Articles are then sampled uniformly from each length cluster.

\subsection{Generation}
In the second stage, the hallucinated version of the data acquired from the previous stage is produced using some well-known LLMs with appropriate instructions. For this purpose, two closed-source LLMs--GPT-4o and Gemini 2.0 Flash—due to their outstanding performance in Persian, along with one open-source LLM, Llama-3.3-70B.”

One of the crucial aspects of this pipeline is developing a strong instruction for the mentioned LLMs to generate hallucinated responses. This instruction consists of four parts: an overview of the goal, hallucination patterns, a few-shot example for each pattern, and output structures. Figure \ref{fig:generation} demonstrates the instruction structure for the QA task. 

The first part of the instruction provides a detailed description of the definition of the task, inputs, outputs, and expected response. Hallucination patterns are scenarios that explain how LLMs must produce hallucinated answers. The same set of diverse patterns is used for both QA and summarization tasks. Five types of patterns are considered for both tasks: contextual misinterpretation, factual fabrication, specificity distortion, incorrect inference, and unwarranted uncertainty \cite{ji2023survey}. An example per pattern illustrates a pair of correct and corresponding hallucinated versions. The last part of the instruction includes constraints on the output structure, such as response length.

The prepared instructions are fed into the three mentioned LLMs, accompanied by the curated, accurate data during the initial data selection phase. Finally, in this stage, there are three hallucinated candidates for each of the received samples.

\subsection{Filtering Hallucinated Candidates}
The objective of the third stage is to obtain the most believable and challenging hallucinated content among the three candidates. Accordingly, a simple prompt format is used, reflecting how most non-expert users normally engage with language models, favoring straightforward prompts over complex engineering techniques \cite{mishra2022help}. Appendix \ref{app:prompts} illustrates the prompt formats for both QA and summarization tasks, accompanied by example inputs.

When this prompt and each candidate are fed into the verifier GPT-4o, it returns “Y” or “N,” along with their corresponding log probabilities. Log probabilities are utilized because, as shown in \cite{kauf2024log}, they provide a more reliable assessment of semantic plausibility than direct zero-shot prompting, which frequently produces inconsistent and inferior outcomes.

One of the challenges is that the models’ response log probabilities contain different characters corresponding to our ‘Y’ and ‘N’ labels. Accordingly, log probabilities are extracted for all ‘Y’ and ‘N’ labels and their equivalents using regular expression pattern (as illustrated in Appendix \ref{app:rejex}) to ensure the accuracy of our evaluation. If an output contains no token that matches either of these regexes, the instance is treated as unparseable and excluded from the analysis. To enhance interpretability, a percentage is derived from the log probability by applying the exponential function. To evaluate the candidates’ credibility, a confidence margin is calculated using the following score function:
\begin{equation}
Score = P('Y') - P('N')
\end{equation}
A higher score indicates that the model regards this response as more plausible, with the rise in P(‘Y’) and decline in P(‘N’). Ultimately, the hallucinated candidate with the greatest score, indicating the most reasonable one to the model, is selected as the final answer, which is the most challenging hallucinated content to be recognized by LLMs. This robust selection is then utilized to evaluate the capability of various LLMs in detecting hallucinatory contents in QA and summarization tasks.

Additionally, a post-hoc analysis of the filtering outputs shows that the finally selected hallucinated candidates are well-distributed across source models (Table~\ref{tab:gpt}), suggesting that our selection procedure does not favor any one generator and thus mitigates generator-specific selection bias.

\begin{table}[h]
\centering
\def\arraystretch{1.2}
\resizebox{0.6\linewidth}{!}{
\begin{tabular}{clc}
\hline
\textbf{Task} & \textbf{Model Name} & \textbf{Share (\%)}\\ 
\hline
\multirow{3}{*}{QA} 
& Gemini-2.0-flash & \text{40\%} \\ 
\cline{2-3}
& GPT-4o & \text{31\%} \\ 
\cline{2-3}
& Llama-3.3-70B & \text{29\%} \\ 
\hline
\multirow{3}{*}{TS} 
& Gemini-2.0-flash & \text{38\%} \\ 
\cline{2-3}
& GPT-4o & \text{30\%} \\ 
\cline{2-3}
& Llama-3.3-70B & \text{32\%} \\ 
\hline
\end{tabular}}
\caption{Source-model share of selected hallucinations.}
\label{tab:gpt}
\end{table}

\begin{figure*}[t]
\includegraphics[width=0.75\paperwidth]{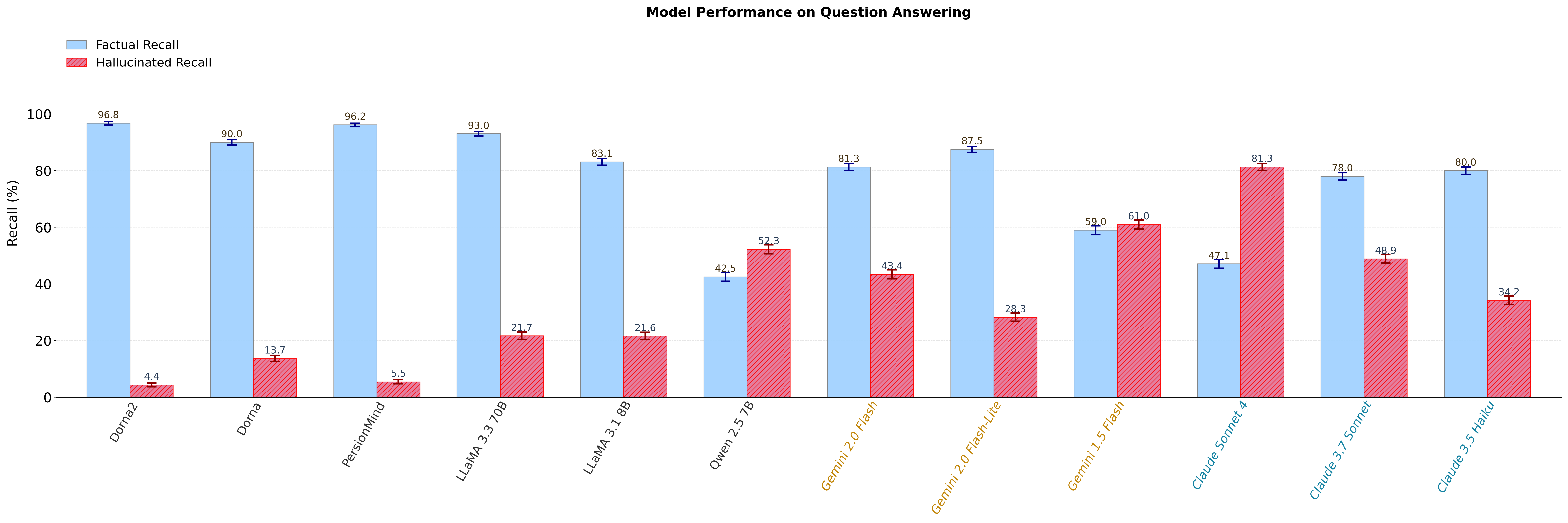}
\caption{Comparison of evaluated LLMs' performances on the QA task.  Error bars show 95\% confidence intervals across evaluation samples. Open-source models are displayed in black, Gemini in orange, and Anthropic in blue.}
\label{fig:QAResult}
\end{figure*}

\subsection{Human Validation}

Besides the above work for the quality and reliability of hallucinated datasets, a full human annotation was also employed. All the 4,000 items in the summarization and QA datasets were manually annotated by three different annotators. As a rule of thumb, in accordance with the instructions provided in Appendix \ref{app:guideline_human}, the item was flagged as hallucination or factual. An item entered the final hallucination set, provided that at least two of the three annotators flagged the item as hallucinated by themselves; else excluded.

This majority-vote decision rule achieved high retention rates in the two datasets. In detail, 3,829 out of 4,000 QA items and 3,917 out of 4,000 summarization items were retained during validation. The inter-annotator agreement was assessed in terms of Gwet’s AC1—reliability index, a measure robust against prevalence and marginal probability bias, and which attained high indices of 0.89 and 0.91 for the QA and summarization set, respectively. These results point to the stable annotation agreement and testify to the validity of the human-curated hallucination labels of the benchmark

\subsection{Evaluation}
\noindent\textbf{Models.} 
In this benchmark, 12 Large Language Models, spanning a wide range of open-source and commercial models, are evaluated. To address Persian-specific performance, some models that are explicitly fine-tuned on Persian datasets are included, specifically Dorna\footnote{https://huggingface.co/PartAI/Dorna-Llama3-8B-Instruct}, Dorna2\footnote{https://huggingface.co/PartAI/Dorna2-Llama3.1-8B-Instruct}, and PersianMind \cite{rostami2024persianmind}. Other multilingual open-source models are Llama 3.1 8B Instruct, Llama 3.1 70B Instruct \cite{grattafiori2024llama3herdmodels}, and Qwen 2.5 7B \cite{yang2024qwen2}. Furthermore, Anthropic's Claude Sonnet 4, Claude Sonnet 3.7, and Claude Haiku 3.5, as well as Google's Gemini family, including Gemini 1.5 Flash, Gemini 2.0 Flash, and Gemini 2.0 Flash-Lite, constitute the evaluated commercial models. For reproducibility, all decoding settings, API/model versions, hardware specifications, and repository links for every evaluated model are documented in Appendix \ref{app:repro}.

\begin{figure*}[t]
\includegraphics[width=0.75\paperwidth]{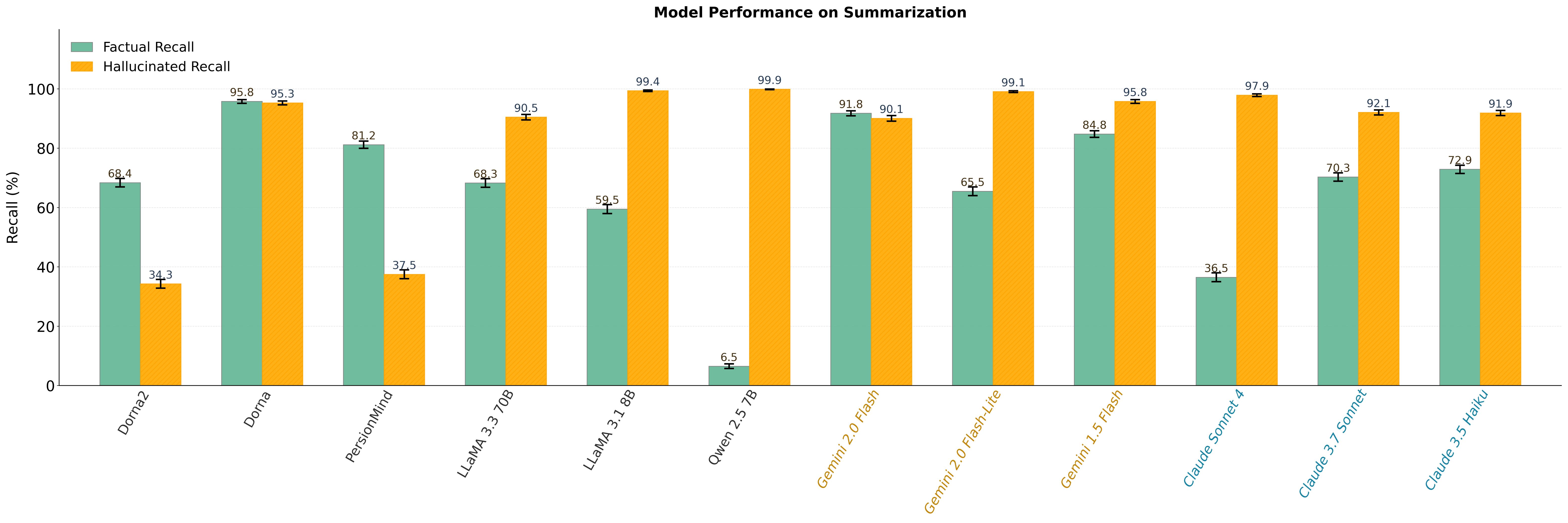}
\caption{Comparison of evaluated LLMs' performances on the summarization task. Error bars show 95\% confidence intervals across evaluation samples. Open-source models are displayed in black, Gemini in orange, and Anthropic in blue.}
\label{fig:SummarizationResult}
\end{figure*}

\vspace{2mm}
\noindent\textbf{Proposed Evaluation Metrics.} To effectively measure model performance in distinguishing between factual (correct) and hallucinated (incorrect) outputs, three targeted measures are introduced for our evaluation: \textit{Factual Recall}, \textit{Hallucinated Recall}, and \textit{Hamming Score}. More info about the Hamming Score can be found in Appendix \ref{app:hamming}. 
\newline
In each evaluation scenario, each prediction made by a model can be categorized into one of four classical outcomes:\newline
    \textbf{True Positive (TP)}: the model correctly identifies a factual output (when the given answer/summary is right, the model also predicts 'Y').\newline
    \textbf{False Positive (FP)}: the model incorrectly predicts hallucinated output as factual (although the given answer/summary is hallucinated, the model predicts 'Y').
    \newline
    \textbf{True Negative (TN)}: the model correctly identifies hallucinated output (when the given answer/summary is hallucinated, the model also predicts 'N').\newline
    \textbf{False Negative (FN)}: the model incorrectly predicts factual output as hallucinated (although the given answer/summary is right, the model predicts 'N').
Based on the above definitions, the metrics are explicitly defined as follows:
\begin{equation}
    \text{Factual Recall} = \frac{TP}{TP + FN}
\end{equation}

\begin{equation}
    \text{Hallucinated Recall} = \frac{TN}{TN + FP}
\end{equation}

The metric \textit{Factual Recall} evaluates the model's reliability in correctly accepting factual information by quantifying the proportion of correctly detected factual outputs among all instances labeled as factual by the dataset. Conversely, the metric \textit{Hallucinated Recall} assesses the capability to precisely detect hallucinated content by measuring the proportion of correctly detected hallucinated outputs within all instances labeled as hallucination in the PerHalluEval.

These two metrics provide complementary evaluations of a model's performance. A high Factual Recall demonstrates the model’s effectiveness in accepting factual content without mistakenly labeling it as hallucinated data, whereas a high hallucinated Recall demonstrates its robustness in identifying hallucinated responses without labeling them as factual data.

\vspace{2mm}
\noindent\textbf{Evaluation Method.} To evaluate the selected LLMs on the benchmark, a straightforward prompting approach is adopted that is consistent with the filtering stage described previously. Specifically, the prompts used for evaluating the models on both QA and summarization tasks are identical to those employed in the filtering step, as illustrated in Figure \ref{fig:filtering}. Each prompt is delivered to the models, starting with defining a role for the LLM with the format \textit{"You are a ..."}. Furthermore, each model is separately provided with both the hallucinated and the correct (non-hallucinated) versions of answers/summaries. The models are instructed to produce a binary classification output ('Y' or 'N') for each provided instance. This evaluation setup enables a systematic and consistent comparison of the models' capabilities in distinguishing hallucinated from accurate content.

\newcommand{\wilson}[3]{#1\%~(95\%~CI: #2--#3)}

\section{Main Results}

\subsection{Question Answering}
Figure~\ref{fig:QAResult} presents the Factual Recall and Hallucination Recall of various LLMs on the PerHalluEval Benchmark for QA. Key findings are explained below.

\vspace{1mm}
\noindent\textbf{Hallucination Recall remains challenging.} Most models struggle to identify hallucinated answers accurately. For example, Gemini 2.0 Flash achieves a Hallucination Recall of only \wilson{43.4}{41.8}{45}, while Llama 3.3 70B reaches \wilson{21.7}{20.4}{23}. This underscores how difficult task it is—even for state-of-the-art models from major companies—to filter out hallucinations reliably.

\vspace{1mm}
\noindent\textbf{Factual Recall and Hallucination Recall are often decoupled.} High Factual Recall does not guarantee strong Hallucination Recall. Claude Sonnet 4 attains moderate Factual Recall (47.1\%) but very high Hallucination Recall (81.3\%), while Dorna2, despite its very high Factual Recall (96.7\%), shows poor Hallucination Recall (4.4\%).

\vspace{1mm}
\noindent\textbf{Large variability among specialized Persian-tuned models.} Persian-specialized models reach high Factual Recall (above 90\%), but Hallucination Recall varies. PersianMind (Factual: \wilson{96.2}{95.6}{96.8}) has low Hallucination Recall (under 6\%), whereas Dorna2 exhibits similar trends, reflecting variability even among strong fine-tuned models on Persian.

\vspace{1mm}
\noindent\textbf{Underperformance in smaller models.} Smaller multilingual models like Qwen-2.5-7B display low Factual Recall (around 40\%) and similarly weak Hallucination Recall, revealing the limitations of compact model architectures in both Factual and Hallucination Recall.

\subsection{Summarization}
Figure~\ref{fig:SummarizationResult} presents the Factual Recall and Hallucination Recall of various LLMs on the PerHalluEval Benchmark for summarization. Key findings are explained below.

\vspace{1mm}
\noindent\textbf{Models exhibit skepticism in accepting summaries.} Several models tend to confidently reject hallucinated summaries (high Hallucination Recall) while being more conservative when accepting correct summaries (lower Factual Recall). For instance, Gemini 2.0 Flash-Lite demonstrates high Hallucination Recall (\wilson{99.1}{98.7}{99.3}) but lower Factual Recall in summary acceptance (\wilson{65.5}{64}{67}).

\vspace{1mm}
\noindent\textbf{High overall Hallucination Recall.} Most models perform strongly on summary hallucination detection. Top performers Qwen 2.5 7B (99.9\%), LLaMA 3.1 8B (99.4\%), Gemini 2.0 Flash-Lite (99.1\%), and Claude Sonnet 4 (97.9\%) demonstrate robust discrimination between faithful and hallucinated summaries.

\vspace{1mm}
\noindent\textbf{Large variability among specialized Persian-tuned models.} Persian-specific models demonstrate divergent results. Dorna achieves both high Factual Recall in summarization (\wilson{95.8}{95.1}{96.4}) and high Hallucination Recall (\wilson{95.3}{94.6}{95.9}). In contrast, PersianMind, despite robust Factual Recall (\wilson{81.2}{79.9}{82.4}), reaches only moderate Hallucination Recall (\wilson{37.5}{36}{39}), highlighting continued variability among targeted Persian-tuned models.

\section{Discussion And Further Analysis}
\subsection{Persian vs. English QA: Minimal Performance Gap}
To evaluate the internal knowledge of LLMs regarding Persian-specific concepts without any external context, this analysis is based on the QA task.
To ensure the quality of the PerHalluEval question answering benchmark, a manual annotation process is conducted. Three annotators reviewed all samples with the guidelines explained in Appendix \ref{app:guideline_persian}, deciding whether the given context and questions were in a Persian context or not. The majority vote (at least two agreeing) decided the final label. Inter-annotator agreement, measured using Fleiss’ kappa, yielded a high score of $\kappa = 0.87$, indicating strong reliability. Finally, 38\% of the PerHalluEval QA benchmark is classified as Persian-specific.

\begin{figure}[h]
    \centering
    \includegraphics[width=\columnwidth]{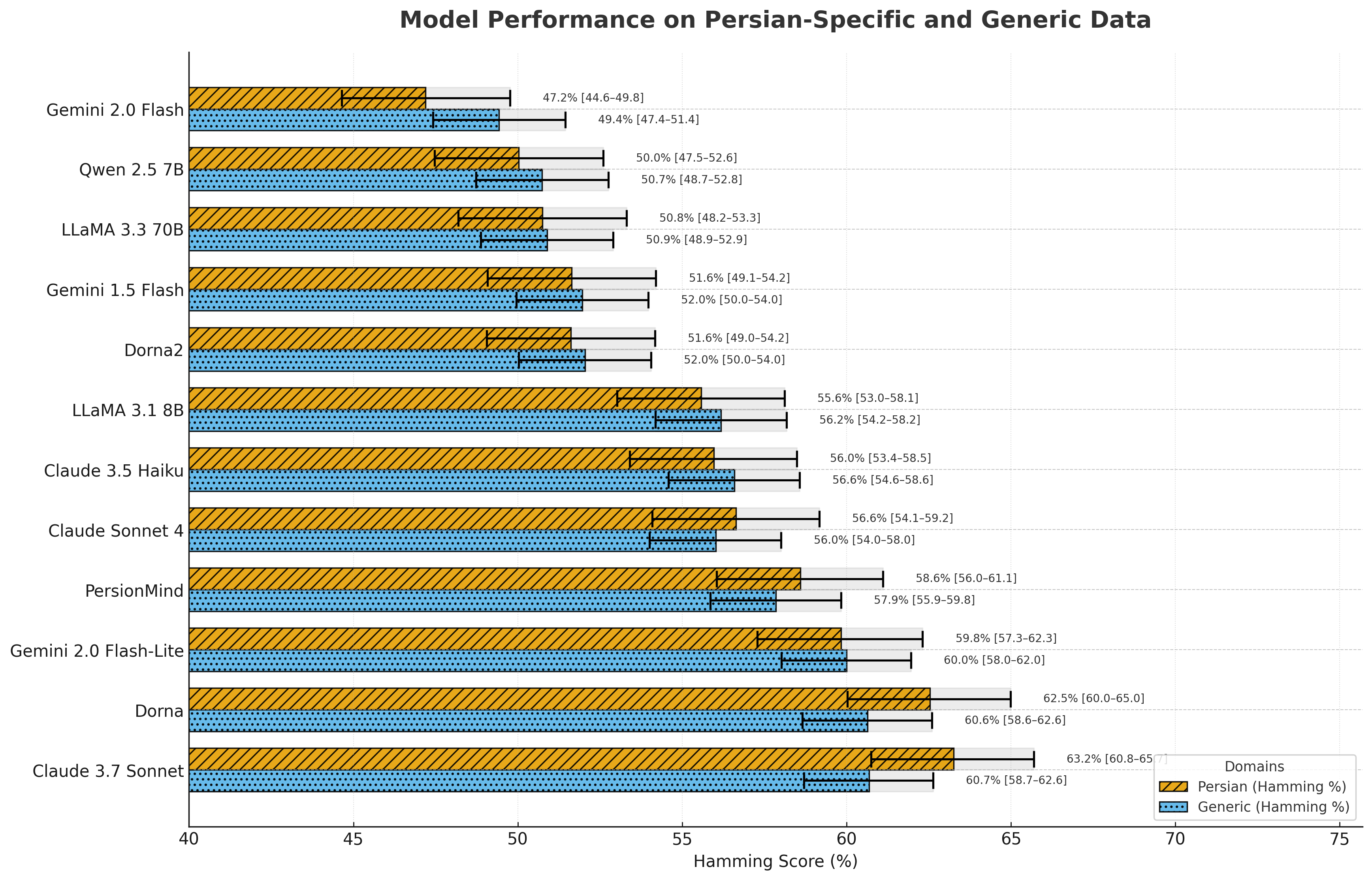}
    \caption{Comparison of LLM performance on Persian-specific vs. non-Persian content in the QA task (Hamming Score).}
    \label{fig:Hamming_persian}
\end{figure}

\begin{table}[h]
\centering
\def\arraystretch{1.2}
\resizebox{\linewidth}{!}{
\begin{tabular}{clccc}
\hline
\textbf{LLM Family} & \textbf{Model Name} & \textbf{Task} & \textbf{Hamming Score} & \textbf{95\% CI} \\
\hline
\multirow{6}{*}{\centering Anthropic}
& \multirow{2}{*}{\centering Claude Sonnet 4}
  & QA & 0.64 & 0.62--0.64 \\
& & TS & 0.67 & 0.66--0.68 \\
\cline{2-5}
& \multirow{2}{*}{\centering Claude Sonnet 3.7}
  & QA & 0.63 & 0.61--0.63 \\
& & TS & 0.81 & 0.80--0.82 \\
\cline{2-5}
& \multirow{2}{*}{\centering Claude Haiku 3.5}
  & QA & 0.57 & 0.55--0.57 \\
& & TS & 0.82 & 0.81--0.83 \\
\hline
\multirow{6}{*}{\centering Gemini}
& \multirow{2}{*}{\centering Gemini 1.5 flash}
  & QA & 0.59 & 0.58--0.60 \\
& & TS & 0.90 & 0.89--0.90 \\
\cline{2-5}
& \multirow{2}{*}{\centering Gemini 2.0 Flash-Lite-preview}
  & QA & 0.57 & 0.56--0.58 \\
& & TS & 0.82 & 0.81--0.83 \\
\cline{2-5}
& \multirow{2}{*}{\centering Gemini 2.0 Flash}
  & QA & 0.62 & 0.60--0.62 \\
& & TS & 0.91 & 0.90--0.91 \\
\hline
\multirow{12}{*}{\centering Open-source}
& \multirow{2}{*}{\centering Qwen 2.5 7B}
  & QA & 0.47 & 0.45--0.48 \\
& & TS & 0.53 & 0.52--0.54 \\
\cline{2-5}
& \multirow{2}{*}{\centering Llama 3.1 8B}
  & QA & 0.52 & 0.51--0.53 \\
& & TS & 0.79 & 0.78--0.80 \\
\cline{2-5}
& \multirow{2}{*}{\centering Llama 3.3 70B}
  & QA & 0.57 & 0.55--0.58 \\
& & TS & 0.83 & 0.82--0.83 \\
\cline{2-5}
& \multirow{2}{*}{\centering Dorna}
  & QA & 0.52 & 0.50--0.52 \\
& & TS & 0.96 & 0.95--0.96 \\
\cline{2-5}
& \multirow{2}{*}{\centering Dorna2}
  & QA & 0.50 & 0.49--0.51 \\
& & TS & 0.51 & 0.50--0.52 \\
\cline{2-5}
& \multirow{2}{*}{\centering PersianMind}
  & QA & 0.51 & 0.49--0.52 \\
& & TS & 0.59 & 0.58--0.60 \\
\hline
\end{tabular}}
\caption{Comparison of evaluated LLMs on QA and text summarization (TS) using Hamming Score with 95\% confidence intervals.}
\label{tab:hamming}
\end{table}

\vspace{1mm}
\noindent\textbf{Model performance on Persian-specific and non-Persian content.} The evaluation results are shown in Figure \ref{fig:Hamming_persian}. Scores cluster in a relatively tight band (roughly mid-40s to low-60s), with modest differences between the two subsets. On the Persian-specific split, the strongest performance is delivered by Claude 3.7 Sonnet, followed closely by Dorna and Gemini 2.0 Flash-Lite. On the generic split, Dorna leads, with Claude 3.7 Sonnet and Gemini 2.0 Flash-Lite close behind. Lower-end performance is observed for Gemini 2.0 Flash and smaller/older open-source models such as Qwen-2.5-7B and LLaMA-3.3-70B.

\vspace{2mm}
\noindent\textbf{Performance gap between Persian-specific and non-Persian content.} Directionally, the gap between Persian-specific and generic QA is small and sometimes mixed: a few models are slightly better on the Persian-specific items, while others favor the generic subset. The overlap of the 95\% confidence intervals indicates that performance differences between Persian and English are not statistically significant and are likely not practically meaningful.

\subsection{The Hamming Olympics: Where Summarization Takes the Gold!}
The comparison between LLMs’ performance on QA and summarization tasks is illustrated in Table \ref{tab:hamming}. One of the most remarkable cross-task observations is that Hamming Scores on summarization are significantly higher than QA scores across almost all assessed models.
This performance disparity can be interpreted in terms of intrinsic and extrinsic hallucinations. In question answering, models depend largely on their internal knowledge, and due to Persian being a low-resource language, the Hamming Score decreases when operating in the Persian context. Conversely, when evaluating LLMs in the summarization task, the original article is provided to the models. This external knowledge guides models to better decide whether the summary contains hallucinated content or not. Two open-source models, Qwen 2.5 7B and Dorna2, are exceptions as they perform well on QA as well as summarization.
Models such as Claude Sonnet 3.7, Gemini 1.5 Flash, Gemini 2 Flash, and Llama 3.3 70B consistently perform at the top across both QA and summarization tasks.

\subsection{The ‘Yes-Man’ Problem: Persian-Tuned LLMs Can’t Say No}
In the QA task, Persian-tuned models (Dorna, Dorna2, and PersianMind) obtained the three lowest performance scores. This observation can be explained by examining the results presented in Figure \ref{fig:QAResult}, in which these models demonstrate the highest Factual Recall. Specifically, the models consistently exhibit a strong inclination toward affirming that the presented answer or content is not hallucinated, regardless of its factual accuracy. Consequently, when assessed using Hallucination Recall, the Persian-tuned models achieve notably lower scores. This disparity indicates a bias within these models toward incorrectly classifying content as accurate and non-hallucinated, thereby resulting in diminished performance on tasks explicitly measuring hallucination sensitivity.

\subsection{A Case Study: Summarization Performance Drop after Persian Fine-Tuning}
Task-dependent behavior is demonstrated by a comparison between Dorna2 and its base model, Llama 3.1 8B.  The models' Hamming Scores in QA are nearly the same, indicating that Persian fine-tuning had a small effect on this task for this particular model pair (see Figure \ref{tab:hamming}). 
In comparison, there is a noticeable difference in the summarization task. Figure \ref{fig:SummarizationResult} shows that Dorna2's Hallucinated Recall is about 60 percentage points lower than Llama 3.1 8B's. Additionally,  the summarization's Hamming Score steadily declines from 0.79 (Llama 3.1 8B) to 0.51 (Dorna2). These findings suggest that in this particular case, summarization seems to be more sensitive to fine-tuning effects.

\section{Conclusion}
In this paper, we introduce PerHalluEval, the first benchmark developed specifically for evaluating hallucination in the Persian language. We propose a three-step automated pipeline to generate appropriate hallucinated samples systematically. In the first stage, suitable samples are extracted from well-known Persian datasets of QA and summarization tasks to analyse extrinsic and intrinsic hallucinations. Subsequently, in the generation stage, three distinct hallucinated candidate outputs are generated for each selected sample, harnessing three powerful LLMs and a strong instruction. The most believable hallucinated candidate is chosen as the final sample after we use a state-of-the-art language model as a judge and log probabilities to evaluate plausibility. Additionally, we invite annotators to specify whether each data point in the QA dataset is specifically related to Persian culture or not for additional investigations.

Our detailed experimental evaluations across 12 diverse LLMs, including commercial, open-source, and specifically fine-tuned models on Persian, show notable limitations in detecting and mitigating hallucination in the Persian language. Our suggested metrics—Hallucination Recall, Factual Recall, and Hamming Score—demonstrate the nuanced challenges these models encounter, highlighting the substantial scope for future advancement in Persian LLMs.

With PerHalluEval, we address a critical gap by introducing a specialized Persian hallucination benchmark to support future NLP research. We believe our benchmark will facilitate further work towards developing trustworthy and culturally-aligned language models for Persian.

\section{Ethical Considerations}
This work creates a benchmark by selecting 4,000 items from pre-existing Persian corpora for QA (PQuAD) and summarization (PN-Summary).  In order to maintain anonymity and confidentiality, three independent undergraduate native Persian annotators received instruction and calibration before labeling; participation was informed and voluntary, and annotator inputs were processed without personal identification.  High agreement and majority vote validation of every item supported dependability while reducing personal burden.

We openly share our findings by providing model identifiers and deterministic assessment parameters to aid in critical review and replication.

\section{Bibliographical References}\label{sec:reference}

\bibliographystyle{lrec2026-natbib}
\bibliography{lrec2026}

\appendix 
\section{Reproducibility and Implementation Details}\label{app:repro}
A deterministic evaluation protocol was adopted. Unless precluded by provider defaults, all systems were evaluated with temperature $=0.0$, top-$p=1.0$, and \texttt{max\_tokens} $=512$. Commercial models were accessed exclusively through their official APIs, and open-source models were executed locally on Google Colab under the indicated GPUs. All open-source models can be found in their respective Hugging Face repositories. Exact identifiers are enumerated in Table~\ref{tab:repro}, which is intended to enable faithful replication of the reported results.

\begin{table*}[t]
\centering
\small
\resizebox{\textwidth}{!}{%
\begin{tabular}{llllccc}
\hline
\textbf{Model} & \textbf{Access} & \textbf{Identifier / Repo} & \textbf{Env./API} & $\mathbf{T}$ & \textbf{top-$p$} & \textbf{Max} \\
\hline
GPT-4o & Closed (API) & \texttt{gpt-4.1-2025-04-14} & OpenAI API & 0.0 & 1.0 & 512 \\
Claude Sonnet 4 & Closed (API) & \texttt{claude-sonnet-4-20250514} & Anthropic API & 0.0 & 1.0 & 512 \\
Claude Sonnet 3.7 & Closed (API) & \texttt{claude-3-7-sonnet-20250219} & Anthropic API & 0.0 & 1.0 & 512 \\
Claude Haiku 3.5 & Closed (API) & \texttt{claude-3-5-haiku-20241022} & Anthropic API & 0.0 & 1.0 & 512 \\
Gemini 1.5 Flash & Closed (API) & \texttt{gemini-1.5-flash} & Google AI Studio & 0.0 & 1.0 & 512 \\
Gemini 2.0 Flash & Closed (API) & \texttt{gemini-2.0-flash} & Google AI Studio & 0.0 & 1.0 & 512 \\
Gemini 2.0 Flash-lite & Closed (API) & \texttt{gemini-2.0-flash-lite} & Google AI Studio & 0.0 & 1.0 & 512 \\
LLaMA 3.1 8B Instruct & Open (Local) & \texttt{meta-llama/Llama-3.1-8B-Instruct} & Colab L4 & 0.0 & 1.0 & 512 \\
LLaMA 3.3 70B Instruct & Open (Local) & \texttt{meta-llama/Llama-3.3-70B-Instruct} & Colab A100 & 0.0 & 1.0 & 512 \\
Qwen 2.5 7B Instruct & Open (Local) & \texttt{Qwen/Qwen2.5-7B-Instruct} & Colab L4 & 0.0 & 1.0 & 512 \\
Dorna & Open (Local) & \texttt{PartAI/Dorna-Llama3-8B-Instruct} & Colab L4 & 0.0 & 1.0 & 512 \\
Dorna2 & Open (Local) & \texttt{PartAI/Dorna2-Llama3.1-8B-Instruct} & Colab L4 & 0.0 & 1.0 & 512 \\
PersianMind & Open (Local) & \texttt{universitytehran/PersianMind-v1.0} & Colab L4 & 0.0 & 1.0 & 512 \\
\hline
\end{tabular}%
}
\caption{Evaluation configuration. Deterministic decoding was used unless a provider’s default prevented overriding a parameter. ``Max'' denotes \texttt{max\_tokens}.}
\label{tab:repro}
\end{table*}

\section{Hamming Score}
\label{app:hamming}
In our model, each input pair generates two binary outputs—one "Factual" and one "Hallucinated"—encoded as a vector $y \in \{0,1\}^2$ with the Factual label $[1, 0]$. Let $f: \mathcal{X} \rightarrow \mathbb{R}^2$ represent our scoring function and $t$ be a thresholding operator (e.g., $t(f_j(x)) = 1$ if $f_j(x) \geq 0.5$ otherwise, $0$). The classifier is defined as $H(x) = t(f(x))$.

We define the multi-label Hamming Loss and Hamming Score as:
\[
L_H(H(x), y) = \frac{1}{2} \sum_{j=1}^{2} \llbracket t(f_j(x)) \neq y_j \rrbracket,
\]
\[
Hamming Score = 1 - L_H(H(x), y),
\]

\tcbset{
  perfectbox/.style={
    enhanced,
    colbacktitle=pastel,
    colback=pastel,
    coltitle=tb-text,  
    colframe=white!20!black,
    fonttitle=\bfseries\LARGE,
    arc=8mm,
    height=19cm,
    boxrule=0.8pt,
    drop shadow southeast={shadow xshift=1mm,shadow yshift=-1mm,opacity=0.3},
    left=8pt, right=8pt, top=8pt, bottom=8pt,
    fonttitle=\bfseries\Large,
    title={#1},
    before skip=8pt, after skip=8pt,
  }
}
\section{Annotation Guidelines}
\subsection{General Guideline for Human Validation}

\label{app:guideline_human}

Annotators were undergraduate native Persian speakers with no overlap with the authorship team. 
They received training examples for each hallucination type and participated in a calibration 
round prior to full annotation.

\begin{tcolorbox}[title=Detailed Annotation Guideline]
You will be shown pairs of text: a gold standard correct version and a potentially hallucinated version 
given by a model. Your task is to determine whether the model's version comprises hallucination(s) in terms of the following types:

\begin{itemize}\setlength\itemsep{2pt}
    \item \textbf{Contextual misinterpretation}: Misunderstanding the context of the source text and coming up with irrelevant or distorted work.
    \item \textbf{Factual fabrication}: Inclusion of matter not in the source and not verifiable.
    \item \textbf{Specificity distortion}: Making the text significantly more general or excessively specific than the source.
    \item \textbf{Wrong assumption}: Drawing a conclusion not warranted by the statement given.
    \item \textbf{Unnecessary doubt}: Suffering from unnecessary doubt or uncertainty about the right facts.
\end{itemize}

\medskip
Mark the sample as \textbf{``Hallucinated''} if any of the above types are represented; otherwise, mark it as \textbf{``Factual.''} 
If you are unsure, choose the label you consider best and leave a brief comment explaining your choice.

\medskip
Focus on the meaning rather than grammar, fluency, or style. Wording differences that do not affect meaning should not be classified as hallucinations. 
If in doubt, re-read the source text carefully and verify that all significant facts remain valid and reliable in the model’s answer.
\end{tcolorbox}

\subsection{Annotating Guideline for Persian-specific and Generic Content}
\label{app:guideline_persian}

Annotators also received explicit instructions to distinguish Persian-specific from generic content.

\begin{tcolorbox}[title=Detailed Guideline for Persian-specific vs Generic Content]
You will be shown Persian-language content. Your task is to determine whether it specifically relates to Persian culture or history, or if it is generic content without unique Persian aspects.

\begin{itemize}\setlength\itemsep{2pt}
    \item Look for clear references to **landmarks, cities, personalities, historical eras, or cultural expressions** associated with countries where Persian is an official language (Iran, Afghanistan, Tajikistan).
    \item Examples of Persian-specific content include mentions of **Persian cities, important historical figures, dynasties, traditional celebrations, or cultural institutions**.
    \item If the content **does not emphasize uniquely Persian elements**, it should be labeled as **generic**.
\end{itemize}

\medskip
Focus on cultural or historical specificity rather than language alone.  
Content written in Persian but describing non-Persian topics should not be considered Persian-specific.
\end{tcolorbox}

\section{Error analysis cases}\label{app:case}
This section presents four representative examples of hallucination errors identified in our evaluation dataset. Each entry includes the question (with English translation when originally in Persian), the hallucinated answer produced by models, the gold (ground-truth) answer, and an analysis of model behavior. These cases highlight different error types that remain challenging for current LLMs.

\definecolor{HallucinatedRed}{HTML}{F6B2B2}
\definecolor{GoldGreen}{HTML}{B7E3B1}
\definecolor{FrameStart}{HTML}{4A90E2} 
\definecolor{FrameEnd}{HTML}{50C9CE}   

\begin{tcolorbox}[
  enhanced,
  colback=white,
  colframe=FrameStart,
  colbacktitle=white,
  coltitle=black,
  title=Factual Fabrication – Numeric Fact,
  boxrule=0.8pt,
  arc=3mm,
  left=4pt,
  right=4pt,
  toprule=1pt,
  bottomrule=1pt,
  borderline west={2pt}{0pt}{FrameStart},
  borderline east={2pt}{0pt}{FrameEnd},
  borderline north={2pt}{0pt}{FrameStart},
  borderline south={2pt}{0pt}{FrameEnd}
]
\textbf{Question:}\\[2pt]
In which year did \textbf{Andi Gutmans} and \textbf{Zeev Suraski} lay the foundations of \textbf{PHP 3}? \\[6pt]

\textbf{Hallucinated Answer:}\\[2pt]
\colorbox{HallucinatedRed!50}{\strut 1997 AD} \\[6pt]

\textbf{Gold Answer:}\\[2pt]
\colorbox{GoldGreen!60}{\strut 1995 AD} \\[8pt]

\textbf{Analysis:}\\[2pt]
\parbox{\linewidth}{Only 1 out of 12 evaluated models (Gemini 1.5 flash) flagged this answer as incorrect. The other 11 accepted the fabricated date, revealing a shared weakness in recalling precise historical numbers.}
\end{tcolorbox}

\begin{tcolorbox}[
  enhanced,
  colback=white,
  colframe=FrameStart,
  colbacktitle=white,
  coltitle=black,
  title=\textbf{Contextual Misinterpretation},
  boxrule=0.8pt,
  arc=3mm,
  left=4pt,
  right=4pt,
  toprule=1pt,
  bottomrule=1pt,
  borderline west={2pt}{0pt}{FrameStart},
  borderline east={2pt}{0pt}{FrameEnd},
  borderline north={2pt}{0pt}{FrameStart},
  borderline south={2pt}{0pt}{FrameEnd}
]
\textbf{Question:}\\[2pt]
What were the demonstrators in \textbf{Bukan} demanding in 1979? \\[6pt]

\textbf{Hallucinated Answer:}\\[2pt]
\colorbox{HallucinatedRed!50}{\parbox{\dimexpr\linewidth-2\fboxsep}{They were demanding the establishment of an Islamic Republic.}} \\[6pt]

\textbf{Gold Answer:}\\[2pt]
\colorbox{GoldGreen!60}{\parbox{\dimexpr\linewidth-2\fboxsep}{They were demanding the release of political prisoners from the Bukan region and its surroundings.}} \\[8pt]

\textbf{Analysis:}\\[2pt]
\parbox{\linewidth}{Only 3 models (assessed Gemini models) flagged this hallucinated answer as incorrect. The other 9—including all three Persian-tuned models—accepted it, indicating difficulty grounding answers in historical socio-political contexts.}
\end{tcolorbox}

\begin{tcolorbox}[
  enhanced,
  colback=white,
  colframe=FrameStart,
  colbacktitle=white,
  coltitle=black,
  title=\textbf{Contextual Misinterpretation Driven by an Ezāfe Noun Chain},
  boxrule=0.8pt,
  arc=3mm,
  left=4pt,
  right=4pt,
  toprule=1pt,
  bottomrule=1pt,
  borderline west={2pt}{0pt}{FrameStart},
  borderline east={2pt}{0pt}{FrameEnd},
  borderline north={2pt}{0pt}{FrameStart},
  borderline south={2pt}{0pt}{FrameEnd}
]
\textbf{Question:}\\[2pt]
Who collaborated on the restoration of the film \textbf{The Splendor of Persepolis}? \\[6pt]

\textbf{Hallucinated Answer:}\\[2pt]
\colorbox{HallucinatedRed!50}{\parbox{\dimexpr\linewidth-2\fboxsep}{Japanese and Italian experts.}} \\[6pt]

\textbf{Gold Answer:}\\[2pt]
\colorbox{GoldGreen!60}{\parbox{\dimexpr\linewidth-2\fboxsep}{A group of Iranology scholars from leading universities worldwide.}} \\[8pt]

\textbf{Analysis:}\\[2pt]
\parbox{\linewidth}{Only 2 models (Gemini 1.5 flash and Claude Sonnet 3.7) answered correctly. The question included a four-word Persian ezāfe noun chain encoding a single proper name (“The Splendor of Persepolis film”). Most models misparsed it, treating only “restoration” as the head and thus accepting a generic hallucinated answer.}
\end{tcolorbox}

\begin{tcolorbox}[
  enhanced,
  colback=white,
  colframe=FrameStart,
  colbacktitle=white,
  coltitle=black,
  title=\textbf{Incorrect Inference},
  boxrule=0.8pt,
  arc=3mm,
  left=4pt,
  right=4pt,
  toprule=1pt,
  bottomrule=1pt,
  borderline west={2pt}{0pt}{FrameStart},
  borderline east={2pt}{0pt}{FrameEnd},
  borderline north={2pt}{0pt}{FrameStart},
  borderline south={2pt}{0pt}{FrameEnd}
]
\textbf{Question:}\\[2pt]
How did \textbf{Shah Abbas} drive the Portuguese out of \textbf{Bandar Abbas}? \\[6pt]

\textbf{Hallucinated Answer:}\\[2pt]
\colorbox{HallucinatedRed!50}{\parbox{\dimexpr\linewidth-2\fboxsep}{By launching a military assault and occupying the port.}} \\[6pt]

\textbf{Gold Answer:}\\[2pt]
\colorbox{GoldGreen!60}{\parbox{\dimexpr\linewidth-2\fboxsep}{By engineering commercial rivalry among European powers on the Persian Gulf coast.}} \\[8pt]

\textbf{Analysis:}\\[2pt]
\parbox{\linewidth}{Only Gemini 2.0 flash detected the hallucinated answer. Although the supporting facts were present in the context, the hallucinated response leapt to an unsupported causal inference (direct warfare) instead of the documented indirect economic strategy described in the gold answer.}
\end{tcolorbox}

\section{Filtering Prompts}
\begin{figure*}[ht]
  \includegraphics[scale=0.34]{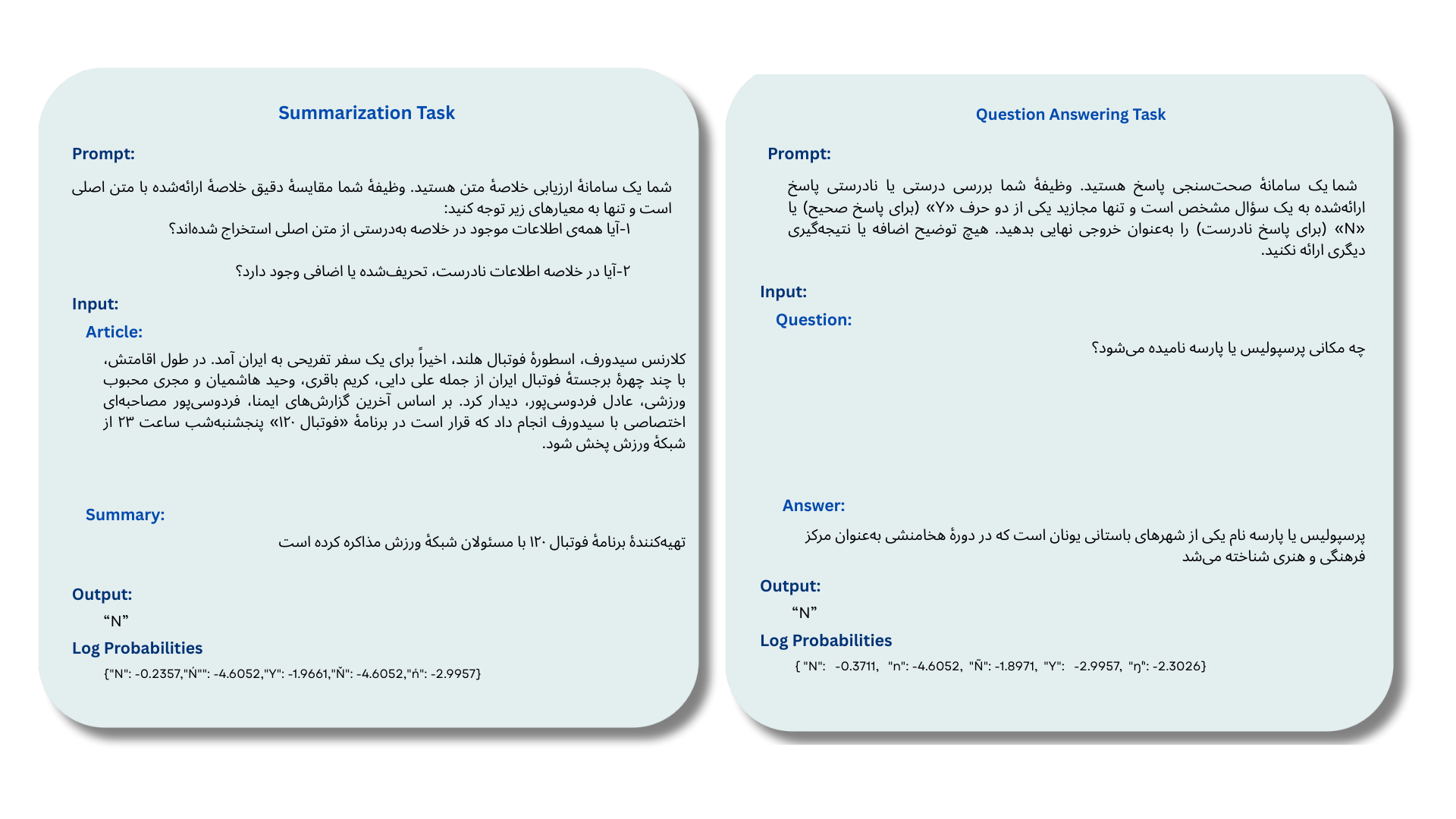}
 \caption{Hallucination filtering guidelines for question-answering and summarization tasks. The image displays the original prompts in Persian.}
\label{fig:fa_filtering}
\end{figure*}

Figures \ref{fig:filtering} and \ref{fig:fa_filtering} are used in the filtering stage to select the best candidate.
\label{app:prompts}

\begin{figure*}[!t]
  \noindent
  \makebox[\textwidth][c]{%
\begin{minipage}[t]{0.48\textwidth}
  \begin{tcolorbox}[perfectbox={Summarization Task}]
    \textbf{Prompt:}\\
    You are a text summary evaluation system. Your task is to accurately compare the provided summary with the original text. When evaluating, focus only on these criteria:
    \begin{enumerate}
      \item Has all the information in the summary been accurately extracted from the original text?
      \item Is there any incorrect, distorted, or additional information in the summary?
    \end{enumerate}

    \medskip
    \textbf{Input:}\\
    \textbf{Article:}\\
    Clarence Seedorf, the Dutch football legend, recently visited Iran for a leisure trip. During his stay, he met with several prominent figures in Iranian football, including Ali Daei, Karim Bagheri, Vahid Hashemian, and the popular sports journalist Adel Ferdosipour. According to the latest reports from IMNA, Ferdosipour conducted an exclusive interview with Seedorf. This interview is set to be broadcast on the “Football 120” program, which airs Thursday night at 11 PM on Iran’s Varzesh (Sports) TV channel.

    \medskip
    \textbf{Summary:}\\
    “The producer of Football 120 has held talks with the officials of the Varzesh (Sports) channel.”

    \medskip
    \textbf{Output:}\\
    “N”

    \medskip
    \textbf{Log Probabilities:}\\
    \texttt{\{
      "N":   -0.2357,
  "Ń": -4.6052,
  "Y":   -1.9661,
  "Ň": -4.6052,
  "ń": -2.9957
    \}
}
  \end{tcolorbox}
\end{minipage}\hfill
\begin{minipage}[t]{0.48\textwidth}
  \begin{tcolorbox}[perfectbox={Question Answering Task}]
    \textbf{Prompt:}\\
    You are a response validity checking system. Your task is to verify whether the provided answer to a specific question is correct or not. You are only allowed to provide one of two letters: ‘Y’ (for correct answer) or ‘N’ (for incorrect answer) as the final output. Do not provide any explanations, extra text, or conclusions. Only give one letter as output: Y or N.

    \medskip
    \textbf{Input:}\\
    \textbf{Question:}\\
    What place was called Persepolis or Parseh?

    \medskip
    \textbf{Answer:}\\
    Persepolis or Parseh is the name of one of the ancient cities of Greece that was known as a cultural and artistic center during the Achaemenid Empire.

    \medskip

    \textbf{Output:}\\
    “N”

    \medskip
    \textbf{Log Probabilities:}\\
    \texttt{\{
"N":   -0.3711,  "Ñ": -4.6052,  "n": -1.8971,  "Y":   -2.9957,  "ŋ": -2.3026
\}
}
    
  \end{tcolorbox}

\end{minipage}
  }
  \caption{Instructions of hallucination filtering for question answering and summarization tasks. The picture shows the English translation of the original Persian prompts.}
  \label{fig:filtering}
\end{figure*}

\section{Precision–Recall Analysis with Iso-F1 Curves}
\label{app:pr_curves}

Insights into model behavior beyond the single scalar metric were sought by plotting the trade-off between precision and recall for all of the models across both tasks (QA and summarization).
Each point was represented as a model, with its position determined by its precision on the x-axis and its recall on the y-axis.
Bubble sizes were scaled proportionally to the F1 scores of the respective models, and gray iso-F1 curves were included as visual guides to aid in the interpretation of the precision/recall trade-off.
Through these graphs, models exhibiting high recall with minimal loss of precision, as well as those biased toward either precision or recall, were revealed.

\begin{figure*}[t]
    \centering
    \begin{minipage}[t]{0.49\textwidth}
        \centering
        \includegraphics[width=\linewidth]{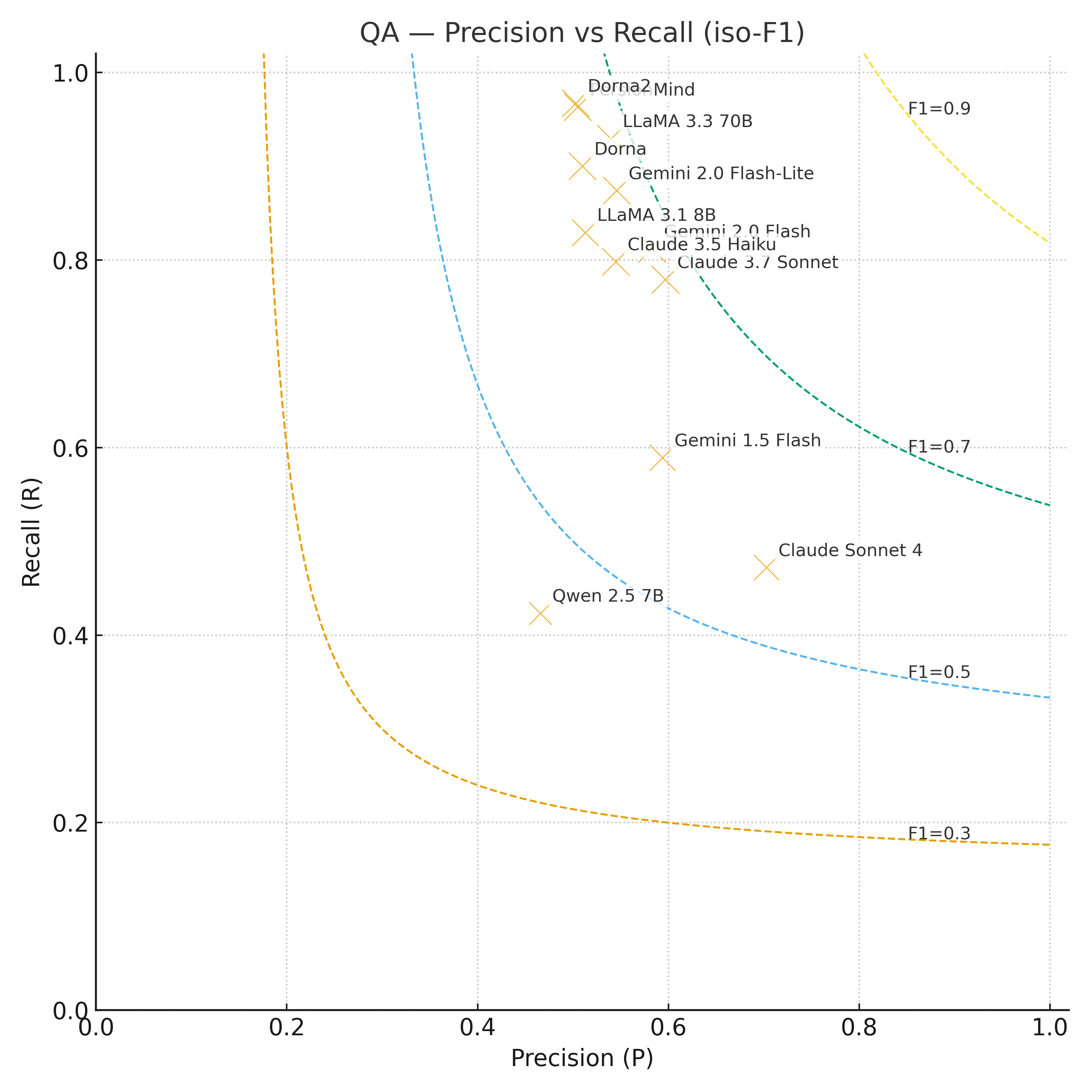}
        \caption*{(a) QA Task}
    \end{minipage}
    \hfill
    \begin{minipage}[t]{0.49\textwidth}
        \centering
        \includegraphics[width=\linewidth]{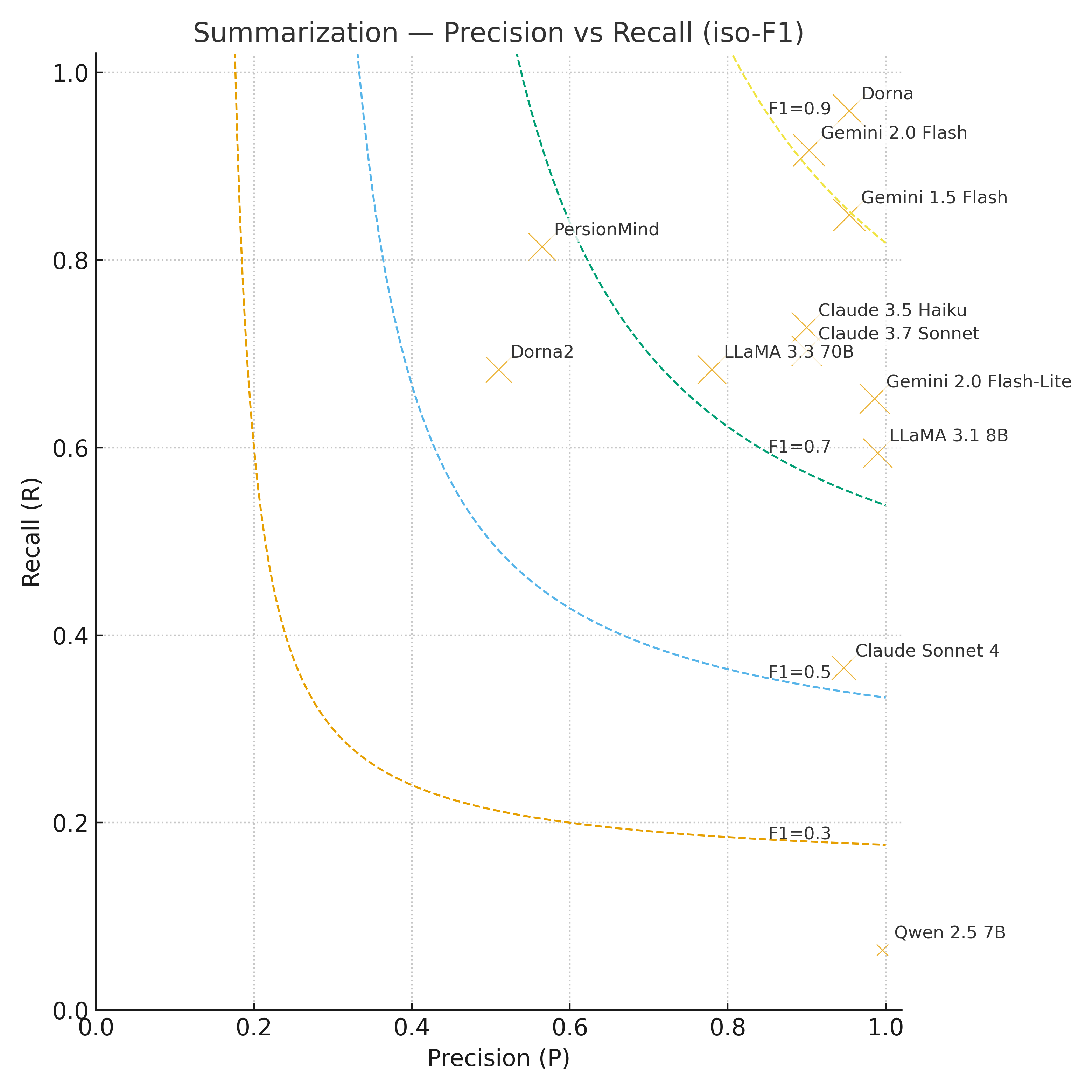}
        \caption*{(b) Summarization Task}
    \end{minipage}
    \caption{Precision–Recall maps with iso-F1 curves. Each point is a model, bubble size reflects its F1 score, and iso-F1 curves (gray) provide guidance on the precision–recall trade-off.}
    \label{fig:pr_curves}
\end{figure*}

\section{Log Probabilities Regex}
\label{app:rejex}
The regular expression for 'Y' is:
\[
\texttt{[ yYỳỲýÝÿŸȳȲẎẏŷŶ¥'.-]+}
\]
\newline
The regular expression for 'N' is:
\[
\texttt{[ nNñÑńŃňŇņŅṇ'.-]+}
\]

\section{Summary of Best Results of Each Model Family}
\label{tab:results}
Table \ref{tab:results} is the table containing a summary of results for QA and summarization with three metrics in each model family.
\begin{table*}[h!]
\centering
\begin{tabular}{| >{\centering\arraybackslash}p{2cm} | >{\centering\arraybackslash}p{1cm} | c | >
{\centering\arraybackslash}p{4cm} | c | c |}
\hline
\textbf{Family} & \textbf{Task} & \textbf{Metric} & \textbf{Best variant} & \textbf{Result} \\
\hline
\multirow{6}{*}{\centering Anthropic} 
& \multirow{3}{*}{\centering QA} 
  & Hamming Score & Claude Sonnet 4 & 0.64 \\
  \cline{3-5}
& 
  & Factual Recall & Claude Haiku 3.5 & 80.0\% \\
  \cline{3-5}
& 
  & Hallucinated Recall & Claude Sonnet 4 & 81.3\% \\
\cline{2-5}
& \multirow{3}{*}{\centering TS} 
  & Hamming Score & Claude Haiku 3.5 & 0.82 \\
  \cline{3-5}
& 
  & Factual Recall & Claude Haiku 3.5 & 72.9\% \\
  \cline{3-5}
& 
  & Hallucinated Recall & Claude Sonnet 4 & 97.9\% \\
\hline
\multirow{6}{*}{\centering Gemini} 
& \multirow{3}{*}{\centering QA} 
  & Hamming Score & Gemini-2.0-flash & 0.61 \\
  \cline{3-5}
& 
  & Factual Recall & Gemini-2.0-flash-lite & 87.5\% \\
  \cline{3-5}
& 
  & Hallucinated Recall & Gemini-1.5-flash & 61.0\% \\
\cline{2-5}
& \multirow{3}{*}{\centering TS} 
  & Hamming Score & Gemini-2.0-flash / Gemini-1.5-flash & 0.90 \\
  \cline{3-5}
& 
  & Factual Recall & Gemini-2.0-flash & 91.8\% \\
  \cline{3-5}
& 
  & Hallucinated Recall & Gemini-2.0-flash-lite & 99.1\% \\
\hline
\multirow{6}{*}{\parbox{2cm}{\centering Fine-tuned on Persian}}
& \multirow{3}{*}{\centering QA} 
  & Hamming Score & Dorna & 0.51 \\
  \cline{3-5}
& 
  & Factual Recall & Dorna2 & 96.8\% \\
  \cline{3-5}
& 
  & Hallucinated Recall & Dorna & 13.7\% \\
\cline{2-5}
& \multirow{3}{*}{\centering TS} 
  & Hamming Score & Dorna & 0.95 \\
  \cline{3-5}
& 
  & Factual Recall & Dorna & 95.8\% \\
  \cline{3-5}
& 
  & Hallucinated Recall & Dorna & 95.3\% \\
\hline
\multirow{6}{*}{\parbox{2cm}{\centering Open-Source Multilingual models}}
& \multirow{3}{*}{\centering QA} 
  & Hamming Score & Llama-3.3-70B & 0.56 \\
  \cline{3-5}
& 
  & Factual Recall & Llama-3.3-70B & 93.0\%\\
  \cline{3-5}
& 
  & Hallucinated Recall & Qwen-2.5-7B & 52.3\% \\
\cline{2-5}
& \multirow{3}{*}{\centering TS} 
  & Hamming Score & Llama-3.1-8B / Llama-3.3-70B & 0.79 \\
  \cline{3-5}
& 
  & Factual Recall & Llama-3.3-70B & 68.3\%\\
  \cline{3-5}
& 
  & Hallucinated Recall & Qwen-2.5-7B & 99.9\% \\
\hline
\end{tabular}
\caption{Comparison of model families across QA and summarization tasks. This table shows the best-performing models within each family based on different evaluation metrics (Hamming, Factual Recall, and Hallucinated Recall). Model families include GPT, Gemini, fine-tuned Persian models, and open-source multilingual models.}
\label{tab:merged_best_models}
\end{table*}

\section{Hallucinated content generation prompts}\label{app:hallprompts}
\begin{figure*}[t]
\centering
\includegraphics[width=0.75\paperwidth]{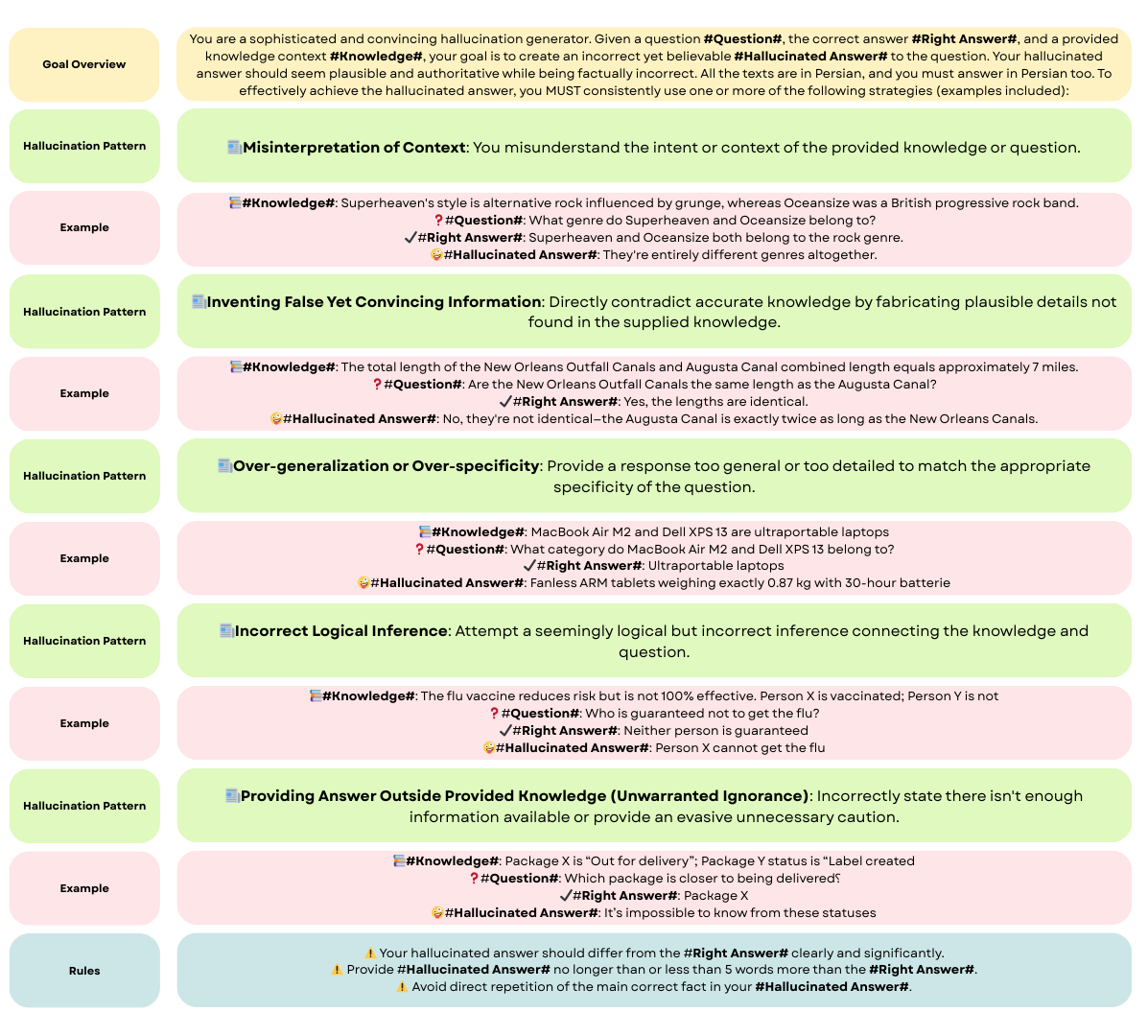}
\caption{The prompt used for QA hallucinated content}
\label{fig:QAPrompt}
\end{figure*}

\begin{figure*}[t]
\centering
\includegraphics[width=0.75\paperwidth]{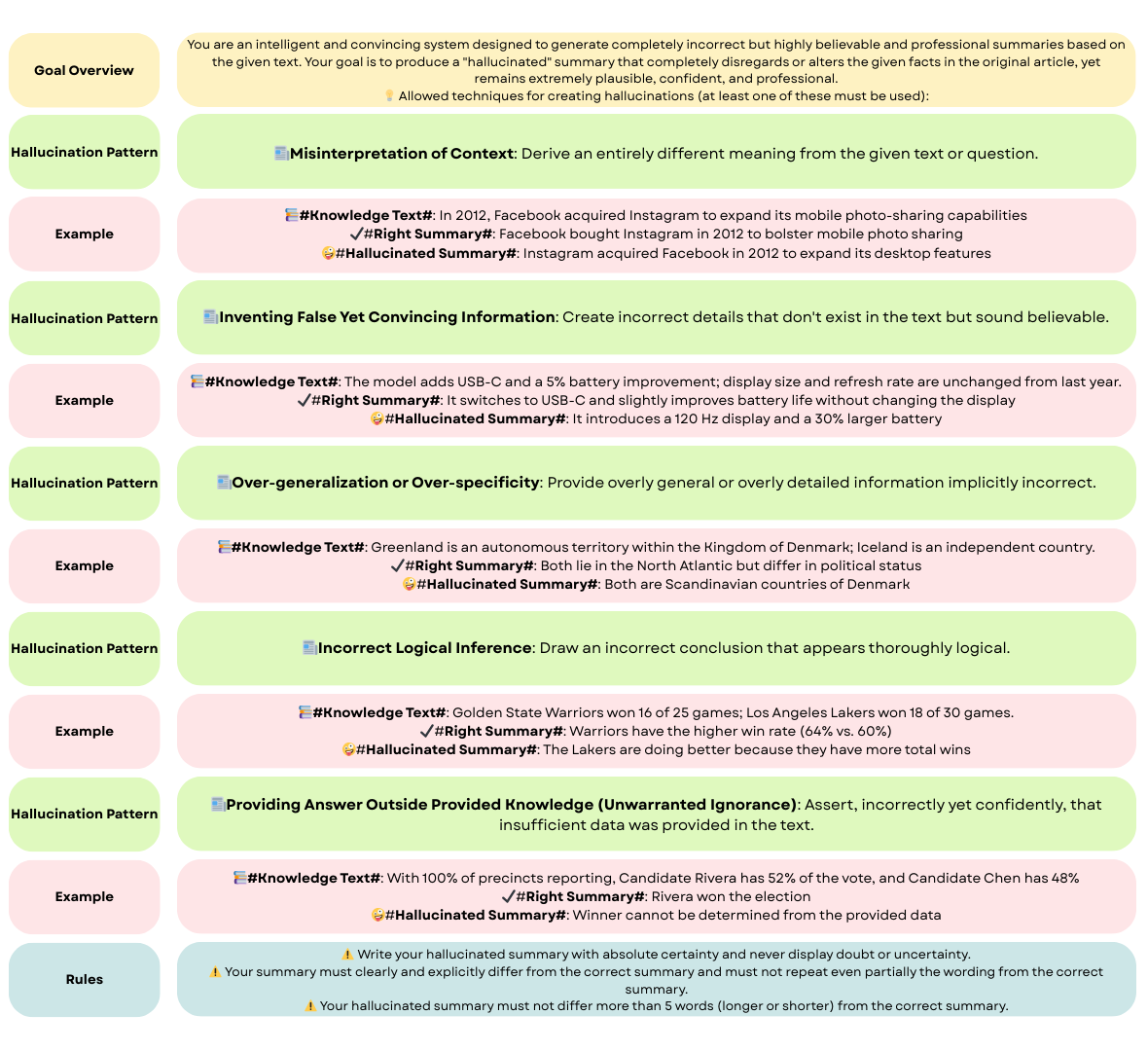}
\caption{The prompt used for Summarization hallucinated content}
\label{fig:SumPrompt}
\end{figure*}

The full prompt for generating hallucinated answers for task QA can be found in Figure \ref{fig:QAPrompt}. The full prompt to generate hallucinated summaries can be found in Figure \ref{fig:SumPrompt}.

\end{document}